\documentclass[lettersize,journal]{IEEEtran}
\usepackage{amsmath,amsfonts}
\usepackage{algorithmic}
\usepackage{algorithm}
\usepackage{array}
\usepackage[caption=false,font=normalsize,labelfont=sf,textfont=sf]{subfig}
\usepackage{textcomp}
\usepackage{stfloats}
\usepackage{url}
\usepackage{verbatim}
\usepackage{graphicx}
\usepackage{cite}
\usepackage[hidelinks,colorlinks=false]{hyperref}
\usepackage{color}
\usepackage{booktabs}

\usepackage{multirow}
\usepackage{diagbox}
\usepackage{scalerel}
\usepackage{tikz}
\usepackage{makecell}
\usetikzlibrary{svg.path}

\definecolor{orcidlogocol}{HTML}{A6CE39}
\tikzset{
  orcidlogo/.pic={
    \fill[orcidlogocol] svg{M256,128c0,70.7-57.3,128-128,128C57.3,256,0,198.7,0,128C0,57.3,57.3,0,128,0C198.7,0,256,57.3,256,128z};
    \fill[white] svg{M86.3,186.2H70.9V79.1h15.4v48.4V186.2z}
                 svg{M108.9,79.1h41.6c39.6,0,57,28.3,57,53.6c0,27.5-21.5,53.6-56.8,53.6h-41.8V79.1z M124.3,172.4h24.5c34.9,0,42.9-26.5,42.9-39.7c0-21.5-13.7-39.7-43.7-39.7h-23.7V172.4z}
                 svg{M88.7,56.8c0,5.5-4.5,10.1-10.1,10.1c-5.6,0-10.1-4.6-10.1-10.1c0-5.6,4.5-10.1,10.1-10.1C84.2,46.7,88.7,51.3,88.7,56.8z};
  }
}

\newcommand\orcidicon[1]{\href{https://orcid.org/#1}{\mbox{\scalerel*{
\begin{tikzpicture}[yscale=-1,transform shape]
\pic{orcidlogo};
\end{tikzpicture}
}{|}}}}

\begin{document}

\title{Cylindrical Mechanical Projector for Omnidirectional Fringe Projection Profilometry}

\author{Mincheol Choi$^{\textsuperscript{\orcidicon{0009-0009-9066-2801}}}$, Gaeun Kim$^{\textsuperscript{\orcidicon{0009-0001-0373-0098}}}$ 
and Jae-Sang Hyun$^{\textsuperscript{\orcidicon{0000-0003-1711-8243}}}$

\thanks{Corresponding author: Jae-Sang Hyun}
\thanks{Mincheol Choi and Jae-Sang Hyun are with the Department of Mechanical Engineering, Yonsei University, Seoul 03722, South Korea. (e-mail: hyun.jaesang@yonsei.ac.kr)}}

\markboth{Journal of \LaTeX\ Class Files,~Vol.~14, No.~8, August~2021}%
{Shell \MakeLowercase{\textit{et al.}}: A Sample Article Using IEEEtran.cls for IEEE Journals}


\IEEEpubid{\begin{minipage}{\textwidth}\ \\[22pt] \centering 
0000--0000/00\$00.00~\copyright~2024 IEEE
\end{minipage}}

\maketitle
\begin{abstract}
The demand for 360-degree 3D reconstruction has significantly increased in recent years across various domains such as the metaverse and 3D telecommunication. Accordingly, the importance of precise and wide-area 3D sensing technology has become emphasized. While the digital fringe projection method has been widely used due to its high accuracy and implementation flexibility, it suffers from fundamental limitations such as unidirectional projection and a restricted available light spectrum. To address these issues, this paper proposes a novel 3D reconstruction method based on a cylindrical mechanical projector. The proposed method consists of a rotational stage and a cylindrical pattern generator with ON/OFF slots at two distinct intervals, enabling omnidirectional projection of multi-frequency phase-shifted fringe patterns. By applying a multi-wavelength unwrapping algorithm and a quasi-calibration technique, the system achieves high-accuracy 3D reconstruction using only a single camera. Experimental results, supported by repeatability and reproducibility analyses together with a measurement uncertainty evaluation, confirm reliable measurement performance and practical feasibility for omnidirectional 3D reconstruction. The expanded uncertainty of the reconstructed depth was evaluated as 0.215 mm.
\end{abstract}



\section{Introduction}

\IEEEPARstart{T}{he} phase-shifting profilometry has advanced into the Digital Fringe Projection (DFP) system, which is widely used in various 3D measurement applications such as robotics, smart manufacturing, entertainment, and even medicine due to its high speed, high accuracy and flexibility of implementation~\cite{li2017high,zhang2018high,hu2023structured}. By integrating with a Digital Light Processing (DLP) projector, DFP enables precise phase-shifting through electronic control, providing a high signal-to-noise ratio and rapid pattern projection~\cite{huang2003high,chen2005miniaturized}. However, since the DLP projector utilizes a silicon-based Digital Micro-mirror Device (DMD), it has limitations of restricted available light spectrum~\cite{dudley2003emerging}. In addition, its inherent unidirectional projection makes it difficult to reconstruct objects from multiple viewpoints, thereby limiting its practicality in modern applications that require omnidirectional measurement.

To resolve these limitations of DFP systems, an alternative projection method based on mechanically generated fringe patterns has been proposed. Heist et al. developed a high-speed structured light system called GOBO (GOes Before Optics) projection~\cite{heist2016high,heist2018gobo}. The GOBO projector generates temporally varying aperiodic sinusoidal fringe patterns using a rotating binary pattern slide. These patterns are captured by two high-speed cameras, and correspondences are identified through normalized cross-correlation of the temporal intensity profiles. This method allows for extremely high-speed 3D measurement, achieving frame rates exceeding 50~kHz. Despite its speed advantages, the inability to determine precise phase shifts limits the achievable measurement accuracy, and since the correspondence is determined based on cross-correlation rather than phase information, the system requires a dual-camera setup. 

On the other hand, Hyun et al. introduced a mechanical projector system for precise phase-based 3D reconstruction~\cite{hyun2018high}. It projects periodic fringe patterns using a rotating wheel with equally spaced slots and captures phase-shifted images via hardware-synchronized high-speed camera and projector. This setup allows accurate phase extraction and, with a disparity refinement algorithm, achieves sub-pixel reconstruction accuracy. Nonetheless, the wheel’s curved surface causes geometric distortion, limiting the effective Field of View (FoV). Additionally, since absolute phase cannot be retrieved, the system still requires two cameras, posing a structural constraint for simplifying the reconstruction setup. To address the structural limitations of previous systems, Liu et al. developed a rotary mechanical projector capable of single-camera 3D reconstruction~\cite{liu2021high}. The system employs a binary-coded grating via error diffusion and optical synchronization to capture high-quality five-step phase-shifted fringe images. The wrapped phase is unwrapped using a reliability map based on intensity modulation, enabling continuous phase recovery. While this unwrapping approach allows reconstruction with only a single camera, its reliance on spatial continuity during phase unwrapping means that it cannot guarantee accurate reconstruction for scenes involving complex geometries or multiple isolated objects.

The aforementioned approaches successfully mitigate the constraints of conventional DFP systems by expanding the usable light spectrum and enabling high-speed 3D measurement. However, these approaches are still based on unidirectional projection, as they employ a planar, disk-shaped rotating wheel whose optical axis is perpendicular to the wheel surface. This geometry inherently causes fan-shaped distortion that becomes more severe toward the outer regions of the projected pattern, resulting in a significantly limited effective FoV. Furthermore, since these mechanical projectors are designed to generate only a single fringe period, absolute phase cannot be recovered, and additional cameras are therefore required to robustly measure isolated features. Such structural constraints increase system cost and implementation complexity, particularly when applied to large-scale or geometrically complex scenes. Choi et al. proposed a metasurface-based structured light system to expand the projection angle beyond the limits of conventional optical components~\cite{choi2024360}. This method enables full 360-degree projection without mechanical rotation by optimizing a metasurface using a differentiable optical model. But the metasurface produces a static structured-light pattern, and the reconstruction relies on a deep-learning-based active stereo framework, which limits accuracy compared with phase-based methods and introduces issues in generalization and computational complexity. Therefore, although it achieves an omnidirectional projection, its reconstruction accuracy and speed remain inferior to those of phase-based approaches.

To overcome the remaining challenges, this paper presents a novel 3D geometry measurement method using a cylindrical-shaped mechanical projector. The proposed projector is designed to project phase-shifted fringe patterns with two distinct spatial periods in all directions by placing ON/OFF slots with different spatial intervals on the surface of a rotating cylinder. A multi-wavelength unwrapping algorithm is employed to retrieve the absolute phase for each pixel from the projected patterns. By adopting quasi-calibration based on the recovered absolute phase, the proposed method simplifies the calibration process and enables high-accuracy 3D reconstruction using only a single camera. To validate its measurement performance, planar and spherical reference targets are evaluated using multiple statistical metrics together with repeatability and reproducibility analyses, and the results are quantitatively compared with a phase-based Stereo Vision (SV) method. Furthermore, a measurement uncertainty analysis based on an uncertainty budget is performed to characterize the measurement performance of the proposed method.

In summary, the key contributions of this paper are:
\begin{enumerate}
    \item {We propose the first cylindrical mechanical projector capable of projecting multi-frequency phase-shifted fringe patterns over $360^\circ$, effectively overcoming the FoV limitations inherent to planar wheel–type mechanical projectors.}

    \item {We design a dual-interval slot structure on the cylindrical pattern generator, enabling multi-wavelength temporal unwrapping and allowing robust absolute phase retrieval even in scenes containing multiple isolated objects.}

    \item {We employ the recovered absolute phase in a quasi-calibration scheme to establish a high accuracy pixel-wise depth estimation framework that operates with only a single camera.}

    \item {We validate the measurement performance of the proposed system through planar and spherical reference experiments, including repeatability and reproducibility analyses, and measurement uncertainty analysis.}
\end{enumerate}

\section{Principle}
\subsection{Cylindrical mechanical projector}
As mentioned in the introduction, conventional DLP projectors have inherent limitations in measurement speed and usable light spectrum due to their operation mechanism. The new type of projector proposed by Hyun utilizes an optical chopper to generate fringe patterns and addressed the limitations of conventional DLP projectors; however, due to its use of a planar optical chopper and a single-period slit configuration, it had a very restricted FoV and could not obtain an absolute phase, requiring at least two cameras for precise reconstruction ~\cite{hyun2018high}. To overcome these challenges, we introduce a cylindrical-shaped mechanical projector that generates an omnidirectional fringe pattern with two different periods. Key differences between prior mechanical projectors and the proposed projector are summarized in Table ~\ref{tab:proj_comparison}.

\subsubsection{Design of the cylindrical-shaped mechanical projector}
\begin{figure}[t]
\centering
\includegraphics[width=3.275in,keepaspectratio]{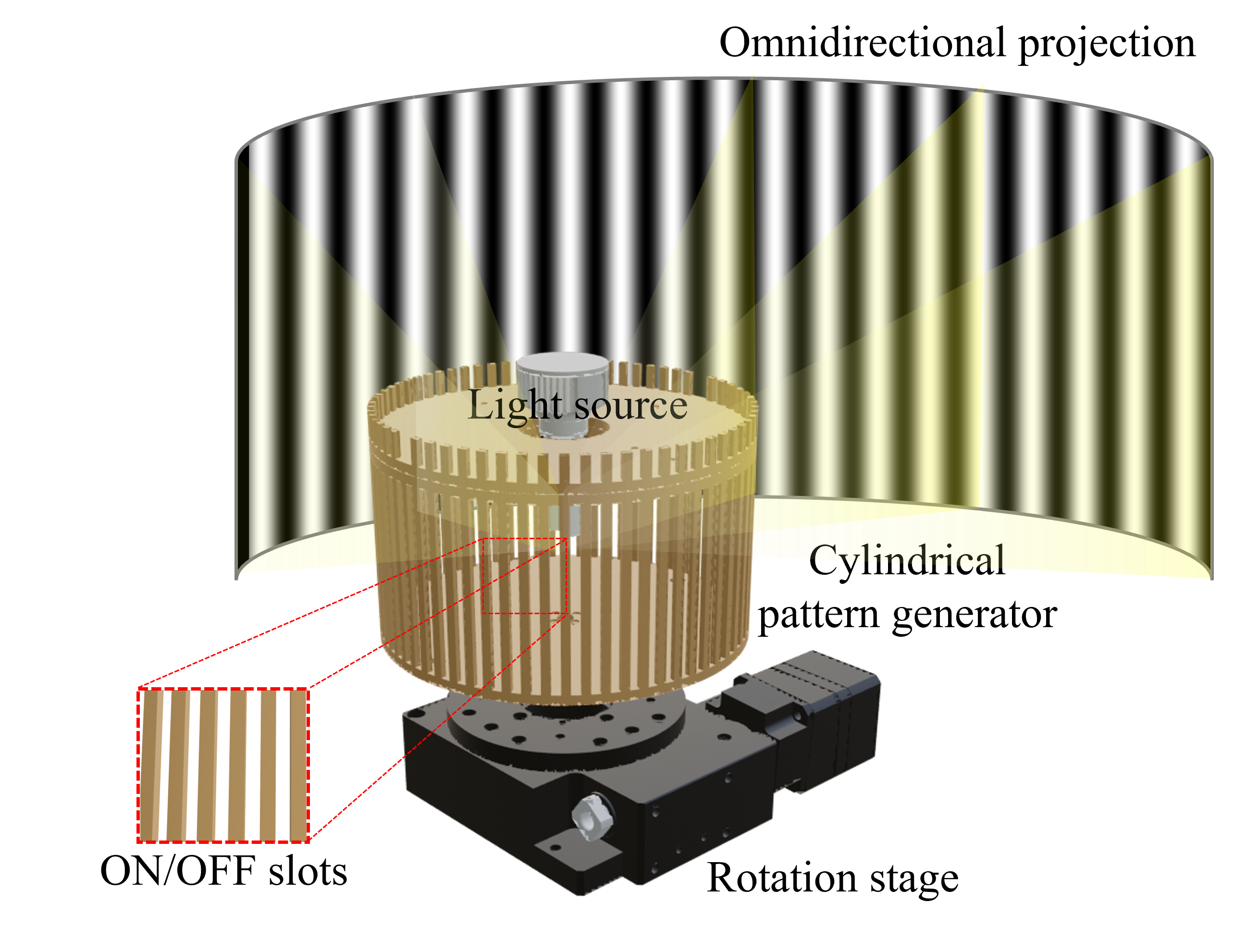}
\caption{Schematic of the proposed cylindrical mechanical projector. 
The red dashed box highlights the ON/OFF slot structure on the cylindrical surface.}
\label{fig_2}
\end{figure}

The pattern generator of the proposed projector was designed in a cylindrical shape to enable the projection of omnidirectional phase-shifted fringe patterns. Fig. \ref{fig_2} illustrates the overall design of the projector. The walls of the pattern generator were configured with evenly spaced ON/OFF slots, ensuring that when a light source is placed at the internal center, fringe patterns are projected outward.

\begin{table}[!t]
\caption{Comparison between prior mechanical projectors and the proposed projector}
\label{tab:proj_comparison}
\centering
\renewcommand{\arraystretch}{1.3}
\begin{tabular}{l|c|c}
\hline
Feature & Prior projectors & Proposed projector  \\
\hline
Pattern generator geometry & Planar wheel & \textbf{Cylindrical} \\
Projection coverage & Unidirectional & \textbf{Omnidirectional} \\
Slit configuration & Single-period & \textbf{Dual-period} \\
Absolute phase retrieval & Limited & \textbf{Enabled} \\
Measurement capability & Relative phase & \textbf{Absolute phase} \\
\hline
\end{tabular}
\end{table}

Using sinusoidal fringe patterns instead of binary patterns is highly advantageous for achieving a more precise phase map with the same number of phase-shifted images~\cite{lei2010digital,li2014some,lei2009flexible}. In DLP projectors, binary patterns are intentionally defocused to generate sinusoidal fringe patterns, improving accuracy. The proposed projector naturally achieves a defocusing effect on binary patterns based on the following principles: 
\textbf{Refraction and Scattering Effects:}  
Light refracts and scatters in various directions depending on the slit thickness and angle of the pattern generator. 
\textbf{Characteristics of the Slit Structure:} As the point light source passes through the slits, it naturally induces diffraction and interference effects. 
\textbf{Influence of the Cylindrical Structure:} Light emitted omnidirectionally from the point light source is reflected and scattered by the inner walls of the cylindrical pattern generator, softening sharp boundaries. Through the combined effects of the above principles, the proposed pattern generator effectively produces fringe patterns with smooth transitions, similar to sinusoidal waves, enabling more accurate phase retrieval.

The fringe pattern generated by the cylindrical pattern generator undergoes distortion depending on the distance. To compensate for these errors, we compute a wrapped phase map using the N-step phase shifting algorithm. The phase shift is determined by Eq.~\eqref{eq_delta}, where $\delta_k$ represents the phase shift value at the $k$-th step, $k$ is the phase-shift time index, and $N$ denotes the total number of phase shifts within one period. In the N-step phase shifting algorithm, a larger $N$ allows for more precise phase retrieval. Therefore, the system was intended to enable fine control of the pattern generator’s shifting using a high-resolution rotation stage.
\begin{align}
\label{eq_delta}
\delta_{k} &= {2\pi{k}}{/N}.
\end{align}

\subsubsection{Design of ON/OFF slots}
For high-accuracy 3D geometry reconstruction of arbitrary multiple objects, obtaining a unique and continuous absolute phase map through temporal unwrapping is essential~\cite{saldner1997temporal, wu2023dynamic}. Most temporal unwrapping methods require multiple pattern images to determine the fringe order~\cite{huntley1993temporal,an2023phase}. Unlike DLP projectors, which can generate various patterns by controlling each pixel of the DMD, prior mechanical projectors rely on a fixed-shape pattern generator with a single-period slit spacing, making temporal unwrapping more challenging.

\begin{figure}[t]
\centering
\includegraphics[width=3.275in,keepaspectratio]{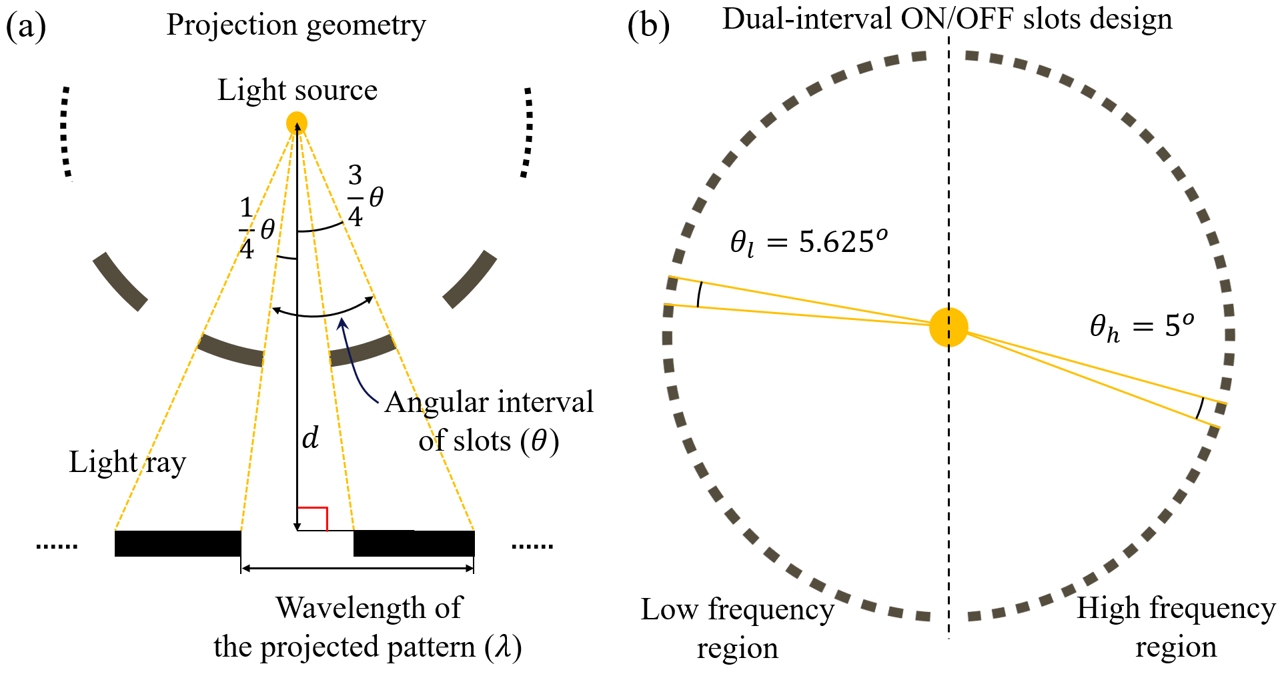}
\caption{Schematic diagram of the pattern generator.
(a) Projection geometry illustrating the relationship between the angular interval $\theta$, the perpendicular distance $d$, and the resulting fringe wavelength $\lambda$ on the projection surface.
(b) Dual-interval ON/OFF slots design on the cylindrical pattern generator, where $\theta_l$ and $\theta_h$ denote the angular intervals corresponding to the low-frequency slot region and the high-frequency slot region, respectively.}
\label{fig_3}
\end{figure}

To address this limitation, we designed the pattern generator with on/off slots having different intervals for every $180^{\circ}$ rotation, as shown in Fig. ~\ref{fig_3}. As the pattern generator rotates, it projects phase-shifted fringe images with two different periods, enabling the implementation of the multi-wavelength unwrapping algorithm for achieving an absolute phase map.

Additionally, since achievable unwrapping range is inversely proportional to the difference between the two fringe periods, it is essential to set the wavelengths appropriately to obtain an absolute phase map across the entire FoV. As illustrated by the projection geometry in Fig.~\ref{fig_3}(a), when the angular interval of slots is sufficiently small, the wavelength of the projected fringe pattern becomes proportional to the interval, which is expressed in Eq.~\eqref{eq_angle}.

\begin{equation}
\label{eq_angle}
\lambda = d\left[\text{tan}(\frac{\theta}{4}) + \text{tan}(\frac{3\theta}{4})\right] \approx d\theta
\end{equation}
Here, $\theta$ denotes the angular interval of slots, $d$ represents the perpendicular distance from the projector to the projection surface, and $\lambda$ is the  wavelength of the projected pattern on the projection surface. Therefore, by setting the intervals to $5^{\circ}$ and $5.625^{\circ}$, the resultant unwrapping coverage range can fully encompass the FoV, allowing the retrieval of an absolute phase map.

\subsection{N-step phase shifting algorithm}
 To retrieve the phase, the N-step phase shifting method is commonly employed due to its advantages, such as high spatial resolution and robustness to noise~\cite{zuo2018phase,he2021comparative}. This method calculates the phase value for each pixel using multiple images, and the fringe pattern with the $k$-th phase shift can be mathematically expressed as follows:
\begin{align}
\label{eq_intensity}
I_{k}(u,v) &= I^{'}(u,v) + I^{''}(u,v) \text{cos}[\phi(u,v) - \delta_{k}],
\end{align}where $I^{'}(u,v)$ is the average intensity, $I^{''}(u,v)$ is intensity modulation, $\phi(u,v)$ is the phase value to retrieve for, and $\delta_{k}$ is the known value of the phase shift.

 To ensure accurate phase calculation, this method compensates for errors and enhances robustness to noise by utilizing additional images. Typically, fringe patterns with more than three images are processed using the least square algorithm, which can be described as

\begin{equation}
\label{eq_lsqr}
    \phi(u,v) =-\text{arctan}\left[\frac{\sum_{k=1}^{N}I_{k}(u,v)\text{sin}\delta_{k}}{\sum_{k=1}^{N}I_{k}(u,v)\text{cos}\delta_{k}}\right].
\end{equation}

Because of the inherent properties of the arctangent function in Eq.~(\ref{eq_lsqr}), the derived phase $\phi(u,v)$ exhibits $2\pi$ discontinuities, ranging from $-\pi$ to $\pi$, and is referred to as the wrapped phase.

\subsection{Multi-wavelength phase unwrapping algorithm}
 To obtain a continuous and unique phase value, a phase unwrapping process is required, which can be expressed as
\begin{equation}
\label{eq_unwrapping}
    \Phi(u,v) = \phi(u,v) + 2\pi \times K(u,v).
\end{equation} 
Here, $\Phi(u,v)$ is the unwrapped phase obtained by resolving the $2\pi$ discontinuities, and $K(u,v)$ represents the fringe order, which is an integer factor determined by the phase unwrapping algorithm. 

\begin{figure}[!t]
\centering
\includegraphics[width=3.275in, keepaspectratio]{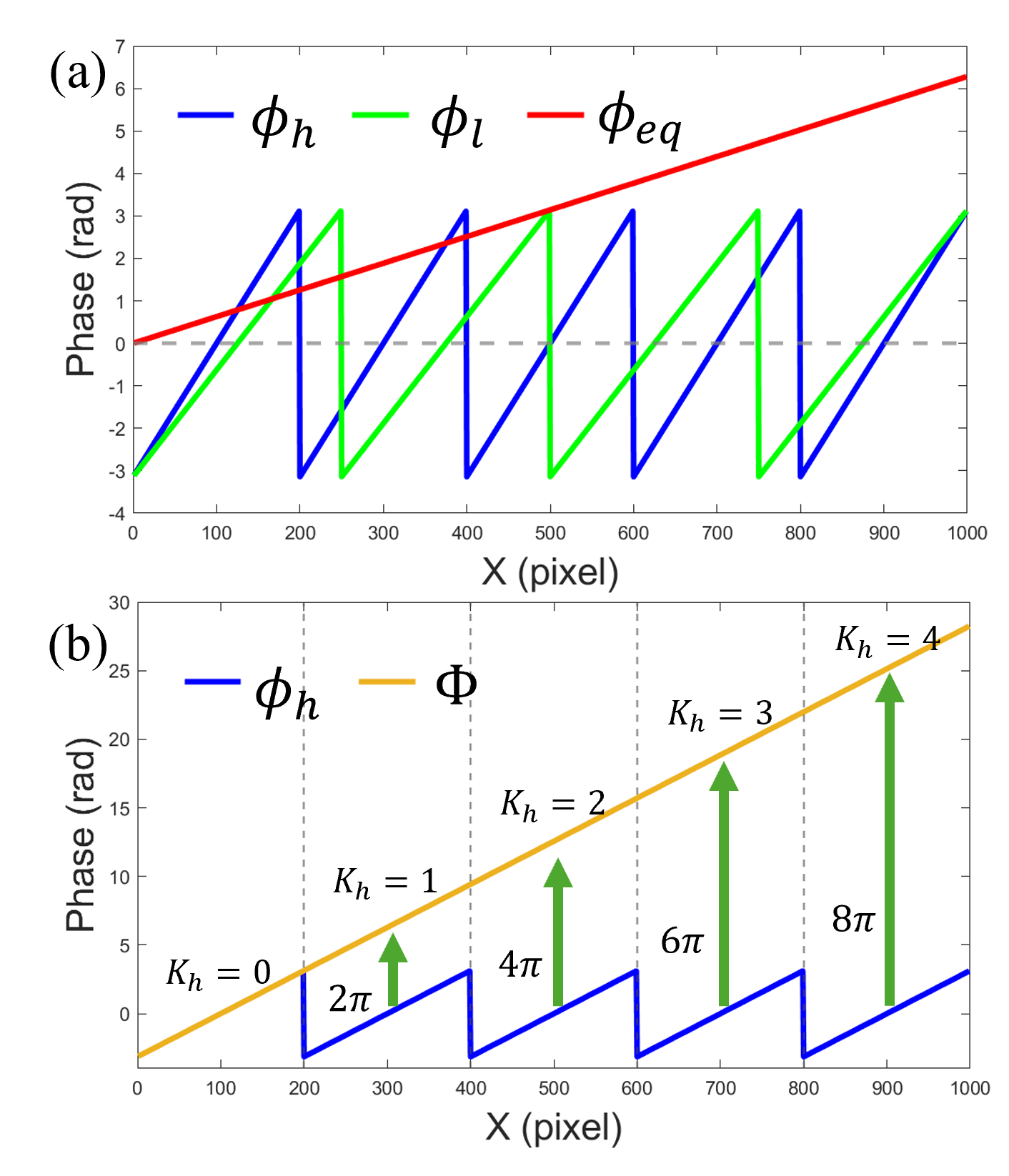}
\caption{Illustration of multiwavelength phase unwrapping process.
A simulation case for $\lambda_h=200$ pixels and $\lambda_l=250$ pixels is shown.
(a) Equivalent phase $\phi_{\mathrm{eq}}$ obtained by combining $\phi_h$ and $\phi_l$, resulting in a continuous phase.
(b) Absolute phase $\Phi$ recovered by adding an integer multiple of $2\pi$ determined by the fringe order.}
\label{fig_1}
\end{figure}

In this work, a multi-wavelength phase unwrapping algorithm is adopted to determine the fringe order using only phase-shifted fringe patterns with two different wavelengths. This approach enables pixel-wise retrieval of the absolute phase without relying on spatial information, allowing reliable phase recovery even in scenes containing isolated objects or discontinuous geometries~\cite{zuo2016temporal,hyun2016enhanced,zhang2018absolute}. The equivalent phase is computed as described in Eq.~\eqref{eq_multi}, where the $\text{Floor}[\cdot]$ function indicates rounding down the value, $\phi_{eq}$ denotes the equivalent phase, and $\phi_h$ and $\phi_l$ denote the wrapped phases obtained from the high-frequency and low-frequency patterns, respectively. Afterward, the subscripts $h$ and $l$ denote high and low frequencies.
\begin{align}
\label{eq_multi}
\phi_{eq}(u,v) &= \phi_{h}(u,v) - \phi_{l}(u,v) \notag \\
&\quad - 2\pi \times \text{Floor}\left [\frac{\phi_{h}(u,v)-\phi_{l}(u,v)}{2\pi} \right]
\end{align}
Although the equivalent phase resolves the phase ambiguity, it becomes more sensitive to noise because it is obtained by subtracting two wrapped phase values ($\phi_h, \phi_l$). Therefore, it is used only as a reference, and the final fringe order is determined as
\begin{align}
\label{eq_order}
K_{h}(u,v) &= \text{Round}\left[\frac{(\lambda_{eq}/\lambda_{h})\phi_{eq}(u,v) - (\phi_{h}(u,v)+\pi)}{2\pi} \right],\\
\label{eq_wavelength}
\lambda_{eq} &= \frac{\lambda_{h}\lambda_{l}}{\lambda_{l}-\lambda_{h}},
\end{align} where the $\text{Round}[\cdot]$ function indicates rounding the value to the nearest integer, $\lambda_{eq}$ denotes the equivalent wavelength, and $\lambda_{h}$ and $\lambda_{l}$ denote the wavelengths corresponding to the high-frequency and low-frequency fringe patterns, respectively. The entire phase retrieval process is depicted in Fig. ~\ref{fig_1}.

Notably, the achievable unwrapping range is equal to the equivalent wavelength. According to Eq. \eqref{eq_wavelength}, a larger gap between the two wavelengths results in a shorter equivalent wavelength. Thus, to achieve unwrapping over the entire FoV, it is important to select wavelengths to adjacent values.

\begin{figure*}[!t]
\centering
\includegraphics[width=6in,keepaspectratio]{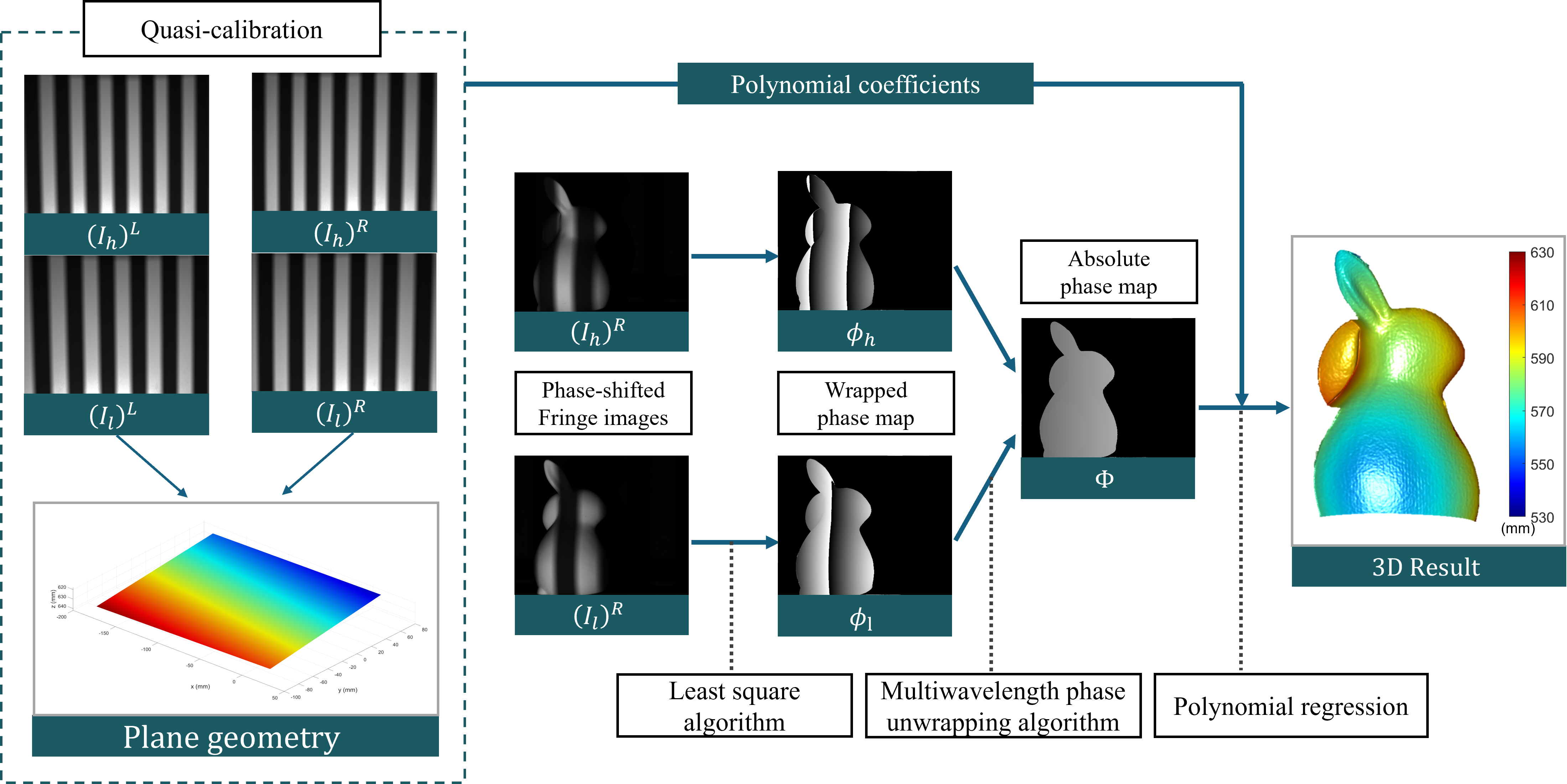}
\caption{Framework of the proposed 3D reconstruction method.
Here, the subscripts $L$ and $R$ denote the left and right cameras, respectively, while $h$ and $l$ indicate the high frequency and low frequency.}
\label{fig_4}
\end{figure*}

\subsection{Quasi-calibration method}
Optical devices such as cameras and DLP projectors are commonly modeled using the pinhole model, which relates the world coordinates $(X,Y,Z)$ to the image plane $(u,v)$ as
\begin{equation}
\label{eq_pinhole}
   s
\begin{bmatrix}
u \\
v \\
1
\end{bmatrix}
=
\mathbf{A}
\begin{bmatrix}
\mathbf{R}&\vert&\mathbf{t}
\end{bmatrix}
\begin{bmatrix}
X \\
Y \\
Z \\
1
\end{bmatrix}
\end{equation}where $s$ is a scaling factor, $\mathbf{A}$ represents the intrinsic matrix, and $[\mathbf{R} \vert \mathbf{t}]$ denotes the extrinsic matrix. Unlike a camera, a projector serves as an output device, which cannot directly obtain its projection image plane, making the calibration process more complicated and time-consuming.

Zhang et al. proposed a camera-projector calibration method that simplifies the projector calibration process by introducing the phase~\cite{zhang2006novel}. This method generates a pseudo-projector image using phase values in horizontal and vertical directions, treating the projector as an inverse camera. By identifying correspondences between the projector and camera images, the pinhole parameters of the projector can be estimated. However, this method requires multiple time-consuming steps to generate the pseudo-projector image and is limited to high-performance DLP projectors capable of controlling patterns in both vertical and horizontal directions.

To overcome these limitations, a new calibration method has been proposed that estimates the relationship between phase and geometry instead of directly estimating pinhole parameters. Previous studies have demonstrated that the relationship between the absolute phase $\Phi$ and the world coordinates $(X, Y, Z)$ can be accurately modeled using a third-order polynomial~\cite{vo2010flexible,huang2014new,vargas2020hybrid,zhang2021flexible}. Mathematically, the polynomial model is expressed as, 
\begin{align}
\label{eq_polynomial}
X(\Phi) &= a_{0} + a_{1}\Phi + a_{2}\Phi^{2} + a_{3}\Phi^{3},\\
Y(\Phi) &= b_{0} + b_{1}\Phi + b_{2}\Phi^{2} + b_{3}\Phi^{3},\\
Z(\Phi) &= c_{0} + c_{1}\Phi + c_{2}\Phi^{2} + c_{3}\Phi^{3},
\end{align} where $a_{0},\cdots, a_{3}$, $b_{0},\cdots, b_{3}$, and $c_{0},\cdots, c_{3}$ are polynomial coefficients obtained through the calibration process.

The quasi-calibration method, which involves measuring a physical planar reference target using a SV-based method and estimating polynomial coefficients from the acquired data, was proposed by Son et al.~\cite{son2024quasi}. In quasi-calibration, the disparity map obtained from the reference plane is refined using phase information, enabling accurate sub-pixel geometry and precise estimation of the polynomial coefficients for each pixel. Because the calibration parameters are derived from measurements of a physical reference target rather than purely numerical assumptions, the resulting phase–height relationship is grounded in experimentally observed geometry. Once calibration is complete, the auxiliary camera is no longer required, and 3D reconstruction can be performed with a single-camera, single-projector system. Since mechanical projectors do not follow the pinhole model, conventional camera-projector calibration methods are not suitable; however, this limitation can be overcome by applying quasi-calibration.

\subsection{Framework of the proposed method}

As shown in Fig. ~\ref{fig_4}, the proposed method follows the steps below to estimate 3D geometric information:

\begin{itemize}
    \item \textbf{Step 1:} Obtain polynomial coefficients through quasi-calibration.
    \item \textbf{Step 2:} Capture phase-shifted fringe image sets with high and low frequencies.
    \item \textbf{Step 3:} Retrieve the absolute phase map by applying the least square algorithm and the multi-wavelength unwrapping algorithm.
    \item \textbf{Step 4:} Calculate the depth information for each pixel by utilizing the polynomial relationship and the absolute phase map.
\end{itemize}

The proposed cylindrical-shaped mechanical projector utilizes a pre-designed, fixed-shape pattern generator; thus, it can only control the pattern in a single direction through rotation. By employing the quasi-calibration method, the limitation of inability to control patterns in both the horizontal and vertical directions can be overcome; moreover, 3D reconstruction with only single-camera, single-projector system is possible. In Step 2, by rotating the pattern generator, phase-shifted fringe image sets with high and low frequencies can be captured. This allows the application of the multi-wavelength phase unwrapping algorithm, enabling the retrieval of the absolute phase for each pixel. Finally, using the absolute phase map and the polynomial relationship obtained in Step 1, high-accurate 3D depth estimation can be performed.

\section{Experiments}
\begin{figure}[!t]
\centering
\includegraphics[width=3.2in, keepaspectratio]{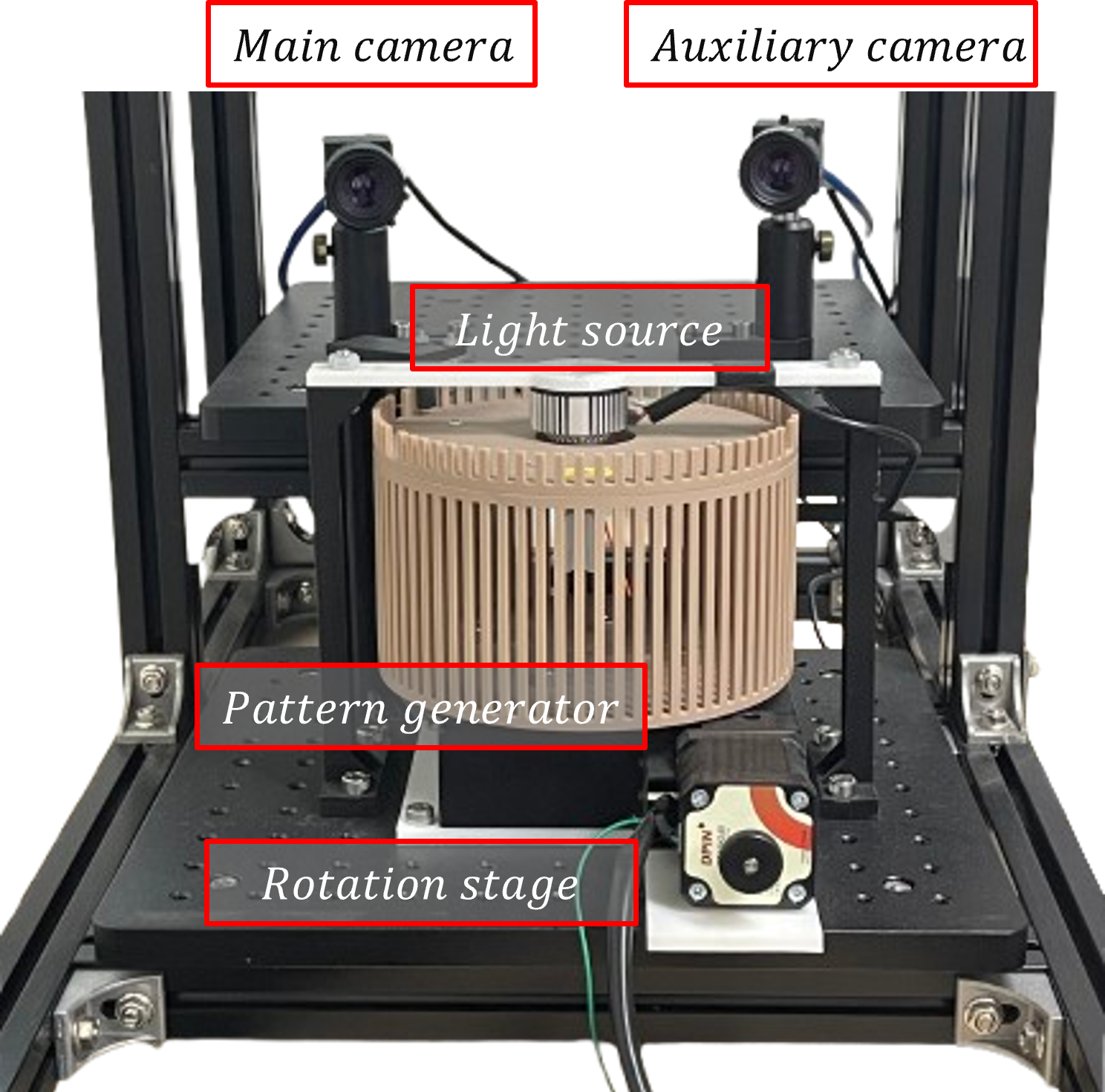}
\caption{Experimental hardware system set up with mechanical projector.}
\label{fig_5}
\end{figure}
\subsection{Experimental setup}

To evaluate the performance of the proposed method, we designed an experimental setup. As shown in Fig. \ref{fig_5}, our setup consists of two CCD cameras (FLIR Blackfly BFS-U3-16S2M) equipped with 12~mm focal length lenses (Computar M1214-MP2), one mechanical projector with a cylindrical-shaped pattern generator, and a rotation stage for phase shifting. The cameras were set to a resolution of 1440 $\times$ 1080, but considering the common FoV of the main and auxiliary cameras, the images were cropped to a resolution of 1000 $\times$ 800.

\begin{figure}[!t]
\centering
\includegraphics[width=3.3in, keepaspectratio]{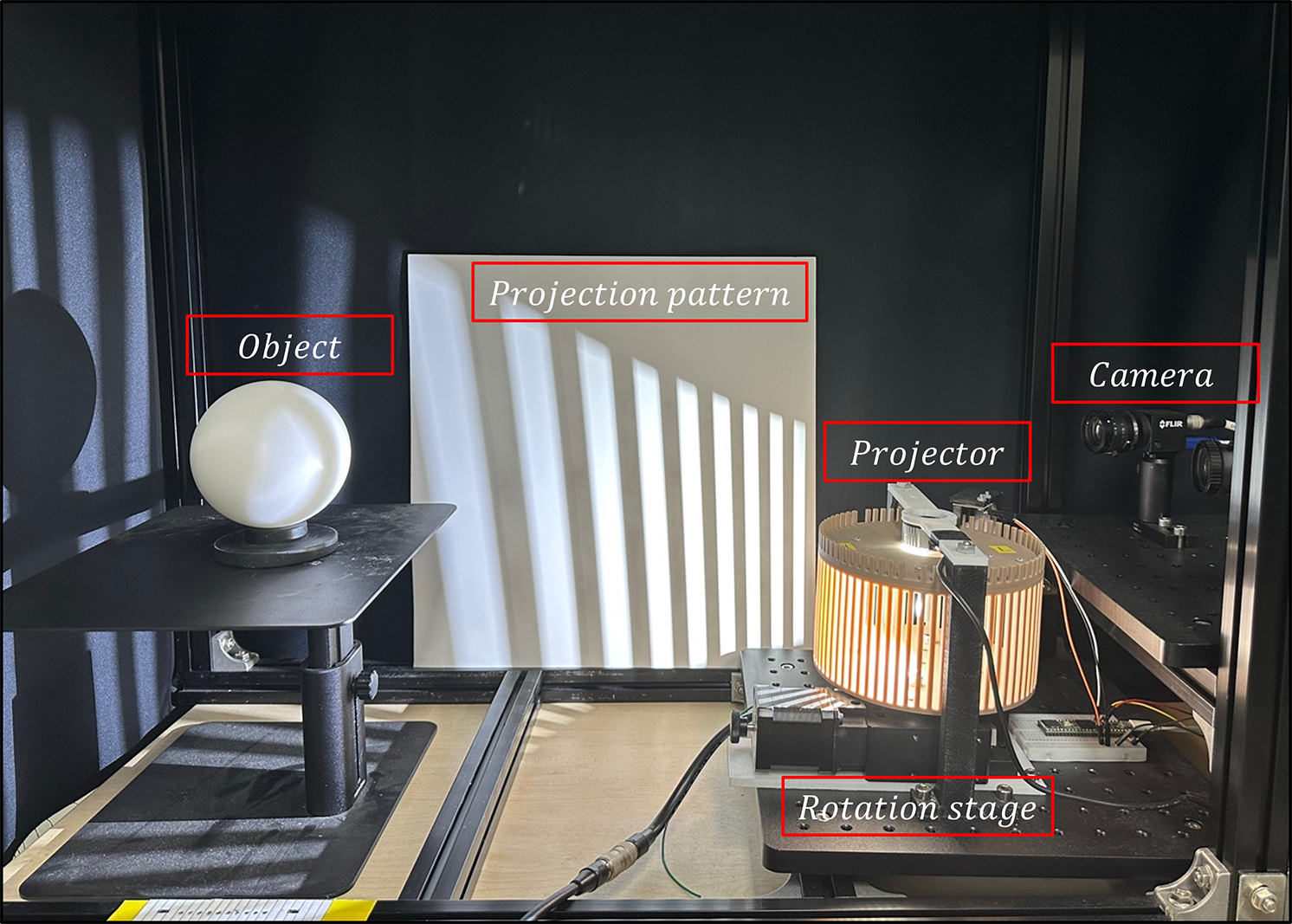}
\caption{Side view of the setup illustrating the fringe pattern projection.}
\label{fig_6}
\end{figure}

Furthermore, as shown in Fig. \ref{fig_6}, due to the projection principle in which light from the central light source passes through the pattern generator to form the fringe pattern, distance-dependent distortion error occurs. Since using more images for phase retrieval aids in compensating for errors caused by distortion, we set the number of phase-shifted images to 25, which is the common divisor of $\theta_h$ and $\theta_l$.

\begin{figure}[!t]
\centering
\includegraphics[width=3.3in, keepaspectratio]{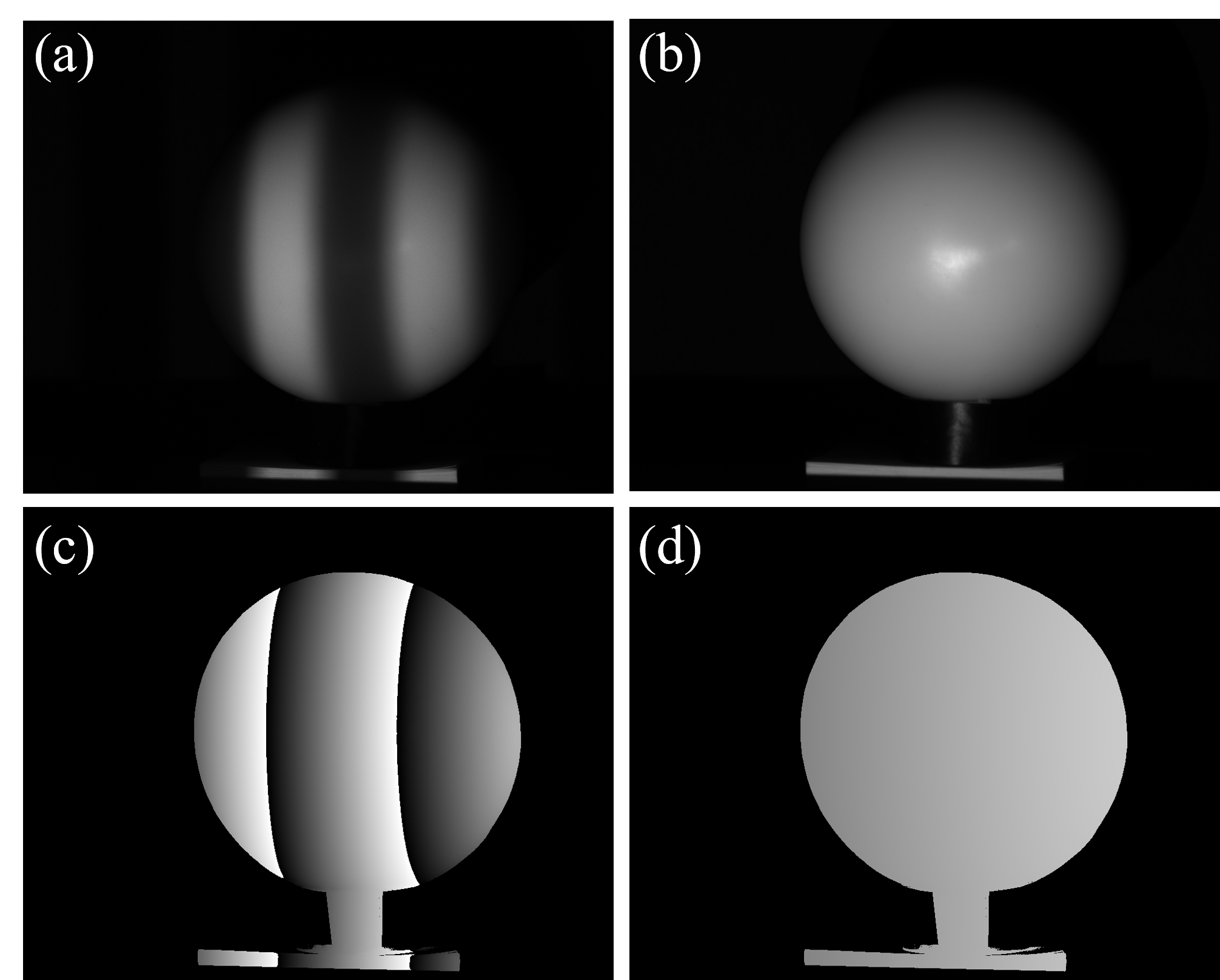}
\caption{Projection result of the proposed method (a) Fringe image from the proposed method, (b) Texture image obtained by integrating 25 phase-shifted fringe images, (c) Wrapped phase map calculated by the least square algorithm, (d) Absolute phase map retrieved using the multi-wavelength unwrapping algorithm.}
\label{fig_7}
\end{figure}

Fig. \ref{fig_7} demonstrates the capability of the proposed mechanical projector to project phase-shifted fringe patterns. Fig.~\ref{fig_7}(a) presents a fringe image acquiring during the experiment. Fig.~\ref{fig_7}(b) shows the texture image, synthesized by combining multiple fringe images. Fig.~\ref{fig_7}(c) illustrates the wrapped phase, obtained by applying the least square algorithm to the fringe images. Fig.~\ref{fig_7}(d) displays the final unwrapped phase, computed by employing the multi-wavelength unwrapping algorithm. The experimental results confirm that the proposed method successfully retrieves the absolute phase for each pixel, which was then utilized in the quasi-calibration process. Previous studies on phase-based fringe projection profilometry with phase–height calibration have reported micrometer-level measurement precision~\cite{hu2020microscopic, liu2003calibration}. For consistency with the achievable measurement precision of phase-based methods, all numerical results in this work are reported up to three decimal places in millimeters.

\subsection{alibration Strategy and Residual Analysis}
\label{sec_calib}
\begin{figure}[!t]
\centering
\includegraphics[width=3.3in, keepaspectratio]{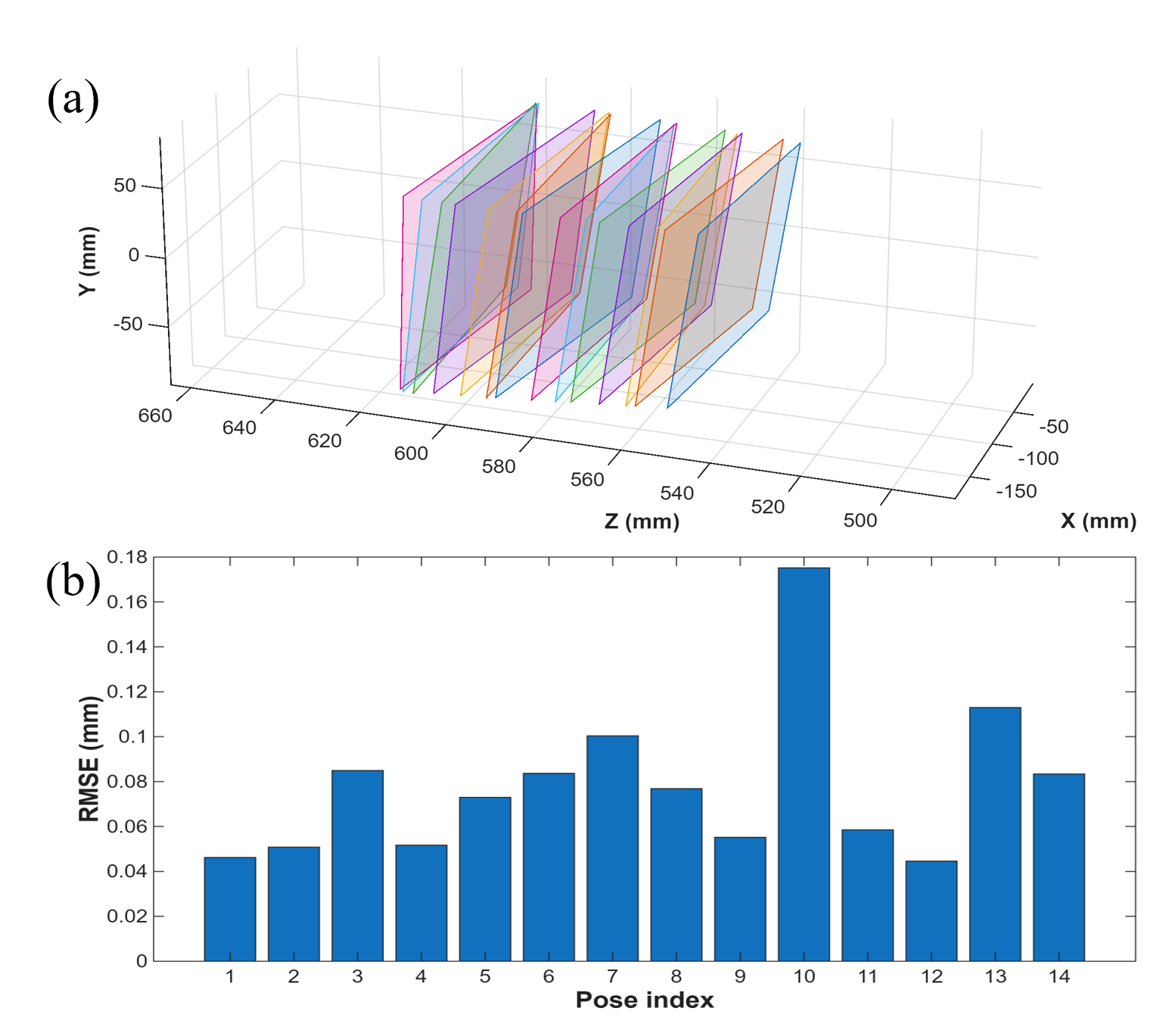}
\caption{Calibration poses distribution and residual evaluation. (a) Spatial distribution of the calibration poses across the measurement volume (540~mm-620~mm), (b) Pose-wise plane residual RMSE for the 14 calibration poses.}
\label{fig_calib}
\end{figure}
A planar calibration board was captured at 64 poses to estimate the intrinsic and extrinsic parameters of the stereo camera pair consisting of the main and auxiliary cameras. After stereo calibration, high- and low-frequency fringe patterns were projected and captured for 14 planar poses within the working depth range from 540~mm to 620~mm, and the absolute phase was computed from the captured fringe images.

For each pose, an initial 3D point set was reconstructed using an SV-based method. In this process, disparity estimation was refined using phase information, enabling sub-pixel-level geometric reconstruction. To obtain a noise-reduced reference geometry of the physical planar calibration board, an ideal plane was fitted to the reconstructed points. Using the phase values and the fitted depth, the third-order polynomial coefficients for the phase-height relationship were estimated. Fig.~\ref{fig_calib}(a) illustrates the spatial distribution of the calibration poses across the measurement volume, where the Z-axis corresponds to the depth direction.

The calibration residual was evaluated as the difference between the depth predicted by the phase–height relationship and the fitted reference plane. The pose-wise Root Mean Square Error (RMSE) was calculated from this residual for each calibration pose. Fig.~\ref{fig_calib}(b) presents the RMSE obtained from the 14 planar poses. The RMSE ranged from 0.045~mm to 0.175~mm, with a median of 0.075~mm. To obtain a global calibration residual, the pooled RMSE was computed using all valid pixels across all calibration poses as
\begin{align}
\label{eq_pooled RMSE}
\sigma_{cal} =
\sqrt{
\frac{1}{P}
\sum_{j=1}^{M}
\sum_{p \in \Omega_j}
r_{p,j}^{2}
}
\end{align} where $r_{p,j}$ denotes the depth residual at pixel $p$ in pose $j$, $M$ represents the number of calibration poses, $\Omega_j$ is the set of valid pixels for pose $j$, and $P$ is the total number of valid pixels. The resulting pooled RMSE was 0.085~mm, which is used as the calibration residual of the system.

\subsection{Plane reconstruction}

To validate the accuracy of the introduced method, we measured a flat plane and compared the results with those obtained by a SV-based method at the same position. 
\begin{figure*}[!t]
\centering
\includegraphics[width=6.735in,keepaspectratio]{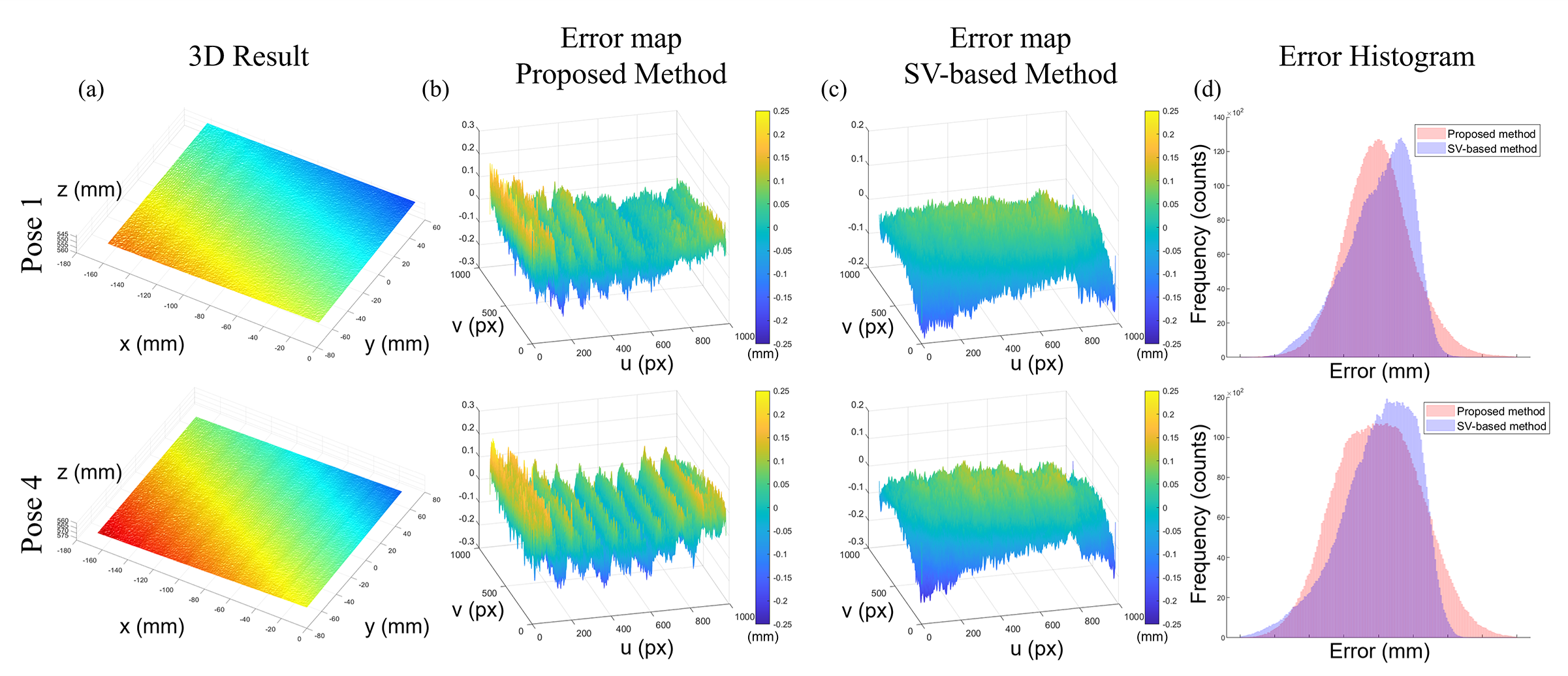 }
\caption{Experimental results of the plane  (a) 3D results from the proposed method, (b) Error maps from the proposed method , (c) Error maps from the SV-based method, (d) Comparison of error histograms between the proposed method and the SV-based method. Each column presents the same type of result for consistent visual comparison across specimens.}
\label{fig_8}
\end{figure*}

Fig. \ref{fig_8} presents the measurement results for the flat plane. Fig.~\ref{fig_8}(a) shows the 3D reconstruction results obtained using the proposed method. To analyze the reconstruction accuracy of the planar surface, an ideal reference plane was estimated from the measured data using the least square method. Based on this reference, the pixel-wise offset errors between the reconstructed surface and the fitted plane were computed. Fig.~\ref{fig_8}(b) and Fig.~\ref{fig_8}(c) illustrate the resulting error maps for the proposed method and the SV-based method, respectively. Fig.~\ref{fig_8}(d) presents overlapping error histograms for the two methods, enabling a direct comparison of their error distributions. Compared to the SV-based method, which shows larger deviations near the boundaries, the proposed method exhibits an error distribution that follows the characteristics of the fringe pattern, indicating the effect of the projected pattern distortion on the measurement accuracy.

\subsubsection{Reproducibility Analysis of Plane Reconstruction}
\renewcommand{\arraystretch}{1.2} 
\begin{table}[!t]
\centering
\caption{Reproducibility results for planar surface reconstruction.}
\label{tab:plane_reproducibility}
\setlength{\tabcolsep}{6pt}
\begin{tabular}{c|cc}
\toprule
\textbf{Pose} 
& \multicolumn{2}{c}{\textbf{RMSE (mm)}} \\
& \textbf{Proposed} & \textbf{SV-based} \\
\midrule
1  & 0.049 & 0.047 \\
2  & 0.061 & 0.051 \\
3  & 0.048 & 0.052 \\
4  & 0.061 & 0.054 \\
5  & 0.039 & 0.055 \\
6  & 0.086 & 0.055 \\
7  & 0.067 & 0.058 \\
8  & 0.081 & 0.065 \\
9  & 0.055 & 0.063 \\
10 & 0.076 & 0.066 \\
\bottomrule
\end{tabular}
\end{table}

In accordance with the ISO GUM framework, repeatability and reproducibility are treated as Type A uncertainty components derived from statistical dispersion across repeated measurements. To quantitatively analyze measurement performance beyond qualitative error characteristics, a reproducibility experiment was therefore carried out. 
Specifically, the planar surface was measured at 10 different poses. 
For each pose, an ideal reference plane was estimated and the reconstruction error with respect to this reference was summarized using the RMSE.

The results are summarized in Table~\ref{tab:plane_reproducibility}. Across the 10 poses, the proposed method achieves a mean RMSE of 0.062~mm, compared to 0.057~mm for the SV-based method, indicating comparable overall reconstruction accuracy under varying measurement configurations. Because the reference plane is estimated from the measured data, the signed residuals exhibit a near-zero mean by construction; therefore, the RMSE primarily represents the spatial dispersion of the residual distribution within each measurement.

The Standard Deviation (STD) of the pose-wise RMSE values is 0.015~mm for the proposed method and 0.006~mm for the SV-based method, respectively. This quantity characterizes the reproducibility of the measurement results by describing the pose-to-pose variability of the reconstruction error. These results indicate that, although the proposed method exhibits higher pose-dependent variability than the SV-based reference, the overall reproducibility remains stable within a sub-0.02~mm range, demonstrating sufficient robustness for planar surface measurement under varying measurement configurations. 

\subsubsection{Repeatability Analysis of Plane Reconstruction}

\renewcommand{\arraystretch}{1.2}
\begin{table}[!t]
\centering
\caption{Repeatability results for planar surface reconstruction.}
\label{tab:comparison_plane_repeat}
\setlength{\tabcolsep}{6pt}
\begin{tabular}{c|cc|cc}
\toprule
\textbf{Repetition} 
& \multicolumn{4}{c}{\textbf{RMSE (mm)}} \\
& \multicolumn{2}{c}{\textbf{Pose 1}} 
& \multicolumn{2}{c}{\textbf{Pose 2}} \\
& \textbf{Proposed} & \textbf{SV-based} 
& \textbf{Proposed} & \textbf{SV-based} \\
\midrule
1 & 0.073 & 0.044 & 0.050 & 0.044 \\
2 & 0.082 & 0.043 & 0.058 & 0.043 \\
3 & 0.085 & 0.044 & 0.056 & 0.044 \\
4 & 0.078 & 0.043 & 0.063 & 0.042 \\
5 & 0.083 & 0.042 & 0.068 & 0.042 \\
\bottomrule
\end{tabular}
\end{table}

To further evaluate the measurement stability under identical conditions, a repeatability analysis was conducted. In this experiment, the planar surface was measured at 2 fixed poses, with 5 repeated acquisitions performed for each pose under identical measurement conditions. For each repetition, the reconstruction error was evaluated using the same statistical metric as those employed in the reproducibility analysis. 

The repeatability results for the planar surface are summarized in Table~\ref{tab:comparison_plane_repeat}. For Pose~1, the proposed method achieved a mean RMSE of 0.080~mm with an STD of 0.005~mm, while the SV-based method yielded a mean RMSE of 0.043~mm with an STD of 0.001~mm. For Pose~2, the proposed method showed a mean RMSE of 0.059~mm with an STD of 0.007~mm, whereas the SV-based method maintained a mean RMSE of 0.043~mm with an STD of 0.001~mm. Here, the STD across repeated acquisitions quantifies the repeatability of the measurement results. These results indicate that the proposed method exhibits slightly higher variability across repeated measurements compared to the SV-based method. Nevertheless, the observed STDs remain below 0.01~mm for both poses, indicating a limited variation across repeated measurements under identical measurement conditions.

\subsection{Sphere reconstruction}
\label{sec_sphere}

We further evaluated the reconstruction performance by comparing the measurement results of a sphere with a 50~mm radius. The reference sphere used in the experiments is a Grade~2 acetal ball. According to the manufacturer’s specification, this corresponds to a diameter tolerance of approximately $\pm 0.051$~mm, which is treated as a Type~B uncertainty associated with the reference artifact. Similarly to the planar surface experiment, the sphere was measured using both the proposed method and the SV-based method. An ideal reference sphere was generated using the radius of 50 mm and the center coordinate estimated from the measured data via the least square method. 

\begin{figure*}[!t]
\centering
\includegraphics[width=6.735in,keepaspectratio]{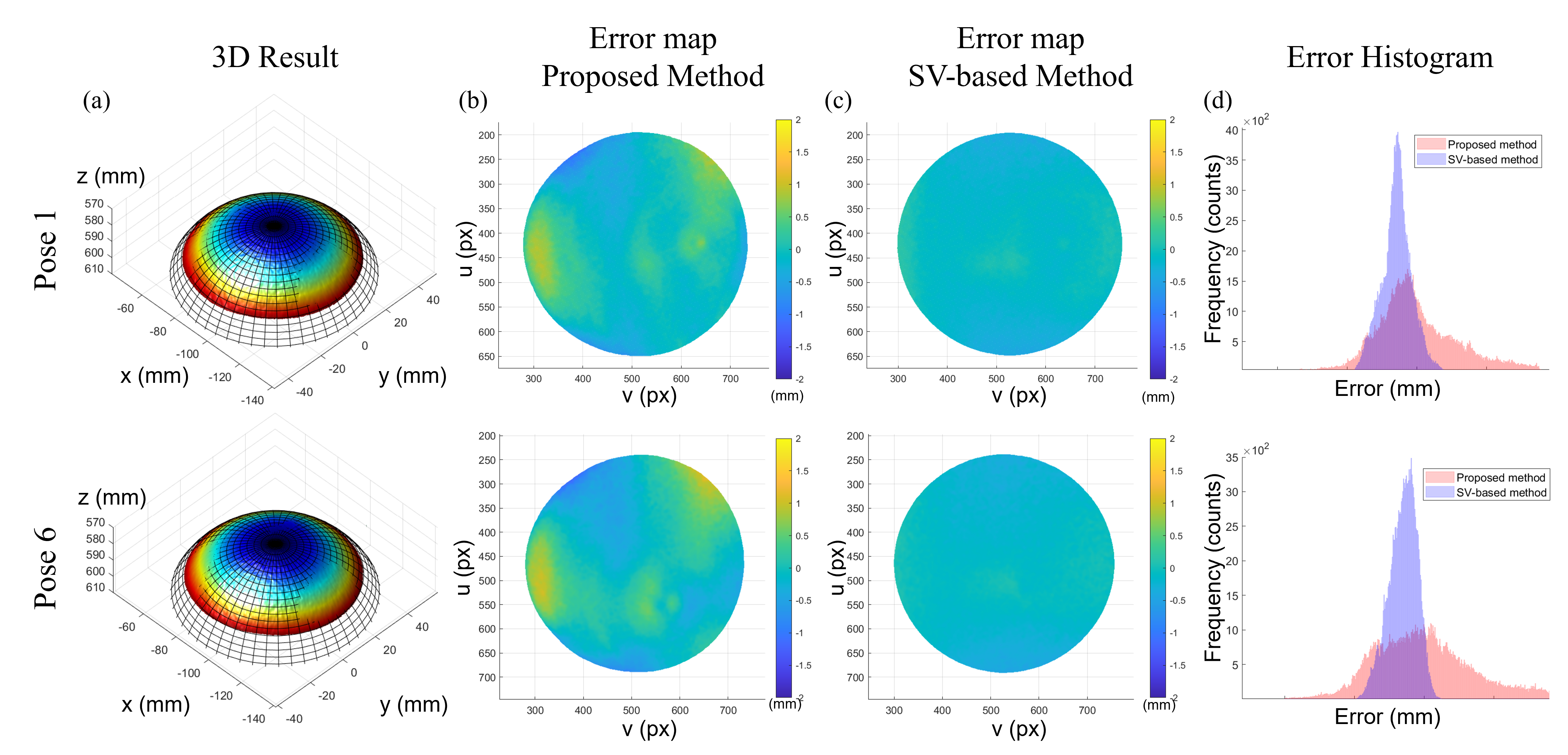}
\caption{Experimental results of the sphere with 50 ~mm radius (a) 3D results from the proposed method, (b) Error maps from the proposed method , (c) Error maps from the SV-based method, (d) Comparison of error histograms between the proposed method and the SV-based method. Each column presents the same type of result for consistent visual comparison across specimens.}
\label{fig_9}
\end{figure*}

Fig.~\ref{fig_9} presents experimental results for the sphere. Fig.~\ref{fig_9}(a) shows the overlay of the reference sphere and the measured data obtained using the proposed method. Fig.~\ref{fig_9}(b) and  Fig.~\ref{fig_9}(c) illustrate the error maps between the measured data and the reference sphere. Fig.~\ref{fig_9}(d) shows overlapping error histograms for the proposed and SV-based methods. Similar to the planar surface results, the proposed method exhibits an error distribution that spatially follows the structure of the projected fringe pattern. The error values are distributed over a wider range, mostly within $\pm$1~mm. In contrast, the SV-based method exhibits a relatively uniform distribution over the entire spherical surface with a narrower error range.

\begin{table*}[!t]
\centering
\caption{Reproducibility results for spherical surface reconstruction.}
\label{tab:sphere_reproducibility}

\setlength{\tabcolsep}{6pt}
\renewcommand{\arraystretch}{1.2}

\begin{tabular}{c|cc|cc|cc|cc}
\toprule
\textbf{Pose}
& \multicolumn{2}{c|}{\textbf{Radius (mm)}}
& \multicolumn{2}{c|}{\textbf{RMSE (mm)}}
& \multicolumn{2}{c|}{\textbf{Mean (mm)}}
& \multicolumn{2}{c}{\textbf{STD (mm)}} \\
& \textbf{Proposed} & \textbf{SV-based}
& \textbf{Proposed} & \textbf{SV-based}
& \textbf{Proposed} & \textbf{SV-based}
& \textbf{Proposed} & \textbf{SV-based} \\
\midrule
1  & 49.994 & 49.880 & 0.303 & 0.181 & -0.003 & -0.146 & 0.303 & 0.107 \\
2  & 49.914 & 49.802 & 0.321 & 0.256 & -0.102 & -0.236 & 0.304 & 0.098 \\
3  & 50.025 & 49.862 & 0.307 & 0.205 & 0.036 & -0.175 & 0.305 & 0.108 \\
4  & 49.950 & 49.812 & 0.407 & 0.252 & -0.056 & -0.227 & 0.403 & 0.110 \\
5  & 50.119 & 49.860 & 0.398 & 0.212 & 0.152 & -0.175 & 0.368 & 0.121 \\
6  & 50.000 & 49.875 & 0.357 & 0.181 & 0.003 & -0.152 & 0.357 & 0.099 \\
7  & 50.022 & 49.832 & 0.425 & 0.235 & 0.033 & -0.205 & 0.424 & 0.114 \\
8  & 49.939 & 49.803 & 0.327 & 0.248 & -0.069 & -0.229 & 0.320 & 0.093 \\
9  & 49.942 & 49.825 & 0.331 & 0.227 & -0.067 & -0.206 & 0.324 & 0.095 \\
10 & 49.988 & 49.846 & 0.282 & 0.196 & -0.012 & -0.178 & 0.282 & 0.081 \\
\bottomrule
\end{tabular}
\end{table*}

\subsubsection{Reproducibility Analysis of Sphere Reconstruction}
\label{subsec:sphere_reproducibility}

To quantitatively compare the measurement performance across different measurement configurations, a reproducibility analysis was conducted for the spherical surface. Specifically, the sphere was measured at 10 different poses. For each pose, an ideal reference sphere was estimated using the same fitting procedure, and the reconstruction error with respect to this reference was characterized using multiple statistical metrics. The estimated sphere radius reflects the geometric accuracy relative to the nominal radius of the reference sphere. The signed mean residual indicates the systematic bias of the reconstruction, while the STD of the residuals describes the spatial dispersion of the reconstruction error within each measurement. The RMSE summarizes the overall reconstruction error by combining both bias and dispersion.

The results are summarized in Table ~\ref{tab:sphere_reproducibility}. Across the 10 poses, the proposed method yielded a mean estimated sphere radius of 49.989~mm with a mean RMSE of 0.346~mm, whereas the SV-based method resulted in a mean radius of 49.840~mm with a mean RMSE of 0.220~mm. For the proposed method, the pose-wise signed mean residual shows an average value of $-0.009$~mm, while the SV-based method exhibits a larger bias of $-0.193$~mm. In addition, the average pose-wise STD of the residuals is 0.339~mm for the proposed method and 0.103~mm for the SV-based method.
The pose-to-pose STD of the estimated radius is 0.059~mm and 0.029~mm for the proposed and SV-based methods, respectively. Similarly, the STD of the pose-wise RMSE values is 0.049~mm and 0.028~mm, and these quantities characterize the reproducibility of the measurement system across different measurement configurations. The observed pose-to-pose variations in the estimated sphere radius and RMSE values suggest that systematic effects contribute significantly to the measurement results, which is consistent with the influence of projection distortion and calibration residuals.

\subsubsection{Repeatability Analysis of Sphere Reconstruction}

\begin{table*}[!t]
\centering
\caption{Repeatability results for spherical surface reconstruction under identical measurement conditions.}
\label{tab:sphere_repeatability}

\setlength{\tabcolsep}{3pt}
\renewcommand{\arraystretch}{1.2}

\begin{tabular}{c|cccc|cccc|cccc}
\toprule
\textbf{Repetition}
& \multicolumn{4}{c|}{\textbf{Radius (mm)}}
& \multicolumn{4}{c|}{\textbf{RMSE (mm)}}
& \multicolumn{4}{c}{\textbf{Mean (mm)}} \\
& \multicolumn{2}{c}{\textbf{Pose 1}} & \multicolumn{2}{c|}{\textbf{Pose 2}}
& \multicolumn{2}{c}{\textbf{Pose 1}} & \multicolumn{2}{c|}{\textbf{Pose 2}}
& \multicolumn{2}{c}{\textbf{Pose 1}} & \multicolumn{2}{c}{\textbf{Pose 2}} \\
& \textbf{Proposed} & \textbf{SV-based}
& \textbf{Proposed} & \textbf{SV-based}
& \textbf{Proposed} & \textbf{SV-based}
& \textbf{Proposed} & \textbf{SV-based}
& \textbf{Proposed} & \textbf{SV-based}
& \textbf{Proposed} & \textbf{SV-based} \\
\midrule
1 & 50.161 & 49.784 & 49.955 & 49.736 & 0.438 & 0.281 & 0.375 & 0.334 & 0.202 & -0.258 & -0.049 & -0.315 \\
2 & 50.185 & 49.784 & 50.013 & 49.767 & 0.454 & 0.280 & 0.394 & 0.300 & 0.231 & -0.258 & 0.021 & -0.277 \\
3 & 50.126 & 49.747 & 49.986 & 49.772 & 0.416 & 0.320 & 0.393 & 0.295 & 0.159 & -0.302 & -0.011 & -0.273 \\
4 & 50.199 & 49.771 & 49.969 & 49.767 & 0.463 & 0.295 & 0.378 & 0.299 & 0.248 & -0.274 & -0.033 & -0.278 \\
5 & 50.177 & 49.758 & 49.959 & 49.746 & 0.448 & 0.310 & 0.391 & 0.323 & 0.221 & -0.289 & -0.044 & -0.303 \\
\bottomrule
\end{tabular}
\end{table*}

To further assess the measurement stability under identical conditions, a repeatability analysis was performed for the spherical surface. In this experiment, the sphere was measured repeatedly at 2 fixed poses, with 5 repeated acquisitions conducted for each pose under identical measurement conditions. For each repetition, the reconstruction error was evaluated using the same statistical metric as those employed in the reproducibility analysis.

The repeatability results for the spherical surface are summarized in Table~\ref{tab:sphere_repeatability}. For Pose~1, the proposed method yields an estimated sphere radius with a mean of 50.170~mm and an STD of 0.028~mm, whereas the SV-based method produces a mean radius of 49.769~mm with an STD of 0.016~mm. The corresponding RMSE values for the proposed method have a mean of 0.444~mm with an STD of 0.018~mm, compared to a mean of 0.297~mm with an STD of 0.018~mm for the SV-based method. The signed mean residual has an average value of 0.212~mm with an STD of 0.034~mm for the proposed method and $-0.276$~mm with an STD of 0.019~mm for the SV-based method.

For Pose~2, the proposed method achieves a mean estimated radius of 49.976~mm with an STD of 0.024~mm, whereas the SV-based method yields a mean radius of 49.758~mm with an STD of 0.016~mm. The mean RMSE for the proposed method is 0.386~mm with an STD of 0.009~mm, compared to 0.310~mm with an STD of 0.017~mm for the SV-based method. The signed mean residual  has an average value of $-0.023$~mm with an STD of 0.029~mm for the proposed method and $-0.289$~mm with an STD of 0.019~mm for the SV-based method.

The repeatability of the measurement system can be quantified by the STD of the RMSE values across repeated measurements. Conservatively, the largest observed STD of the RMSE is 0.018~mm, which is used as the repeatability metric of the system. Overall, the estimated radius, RMSE, and signed mean residual exhibit limited variation across repetitions, indicating stable geometric estimation and bias behavior under identical measurement conditions.

\subsection{Arbitrary surface reconstruction}
\begin{figure}[!t]
\centering
\includegraphics[width=3.3in,keepaspectratio]{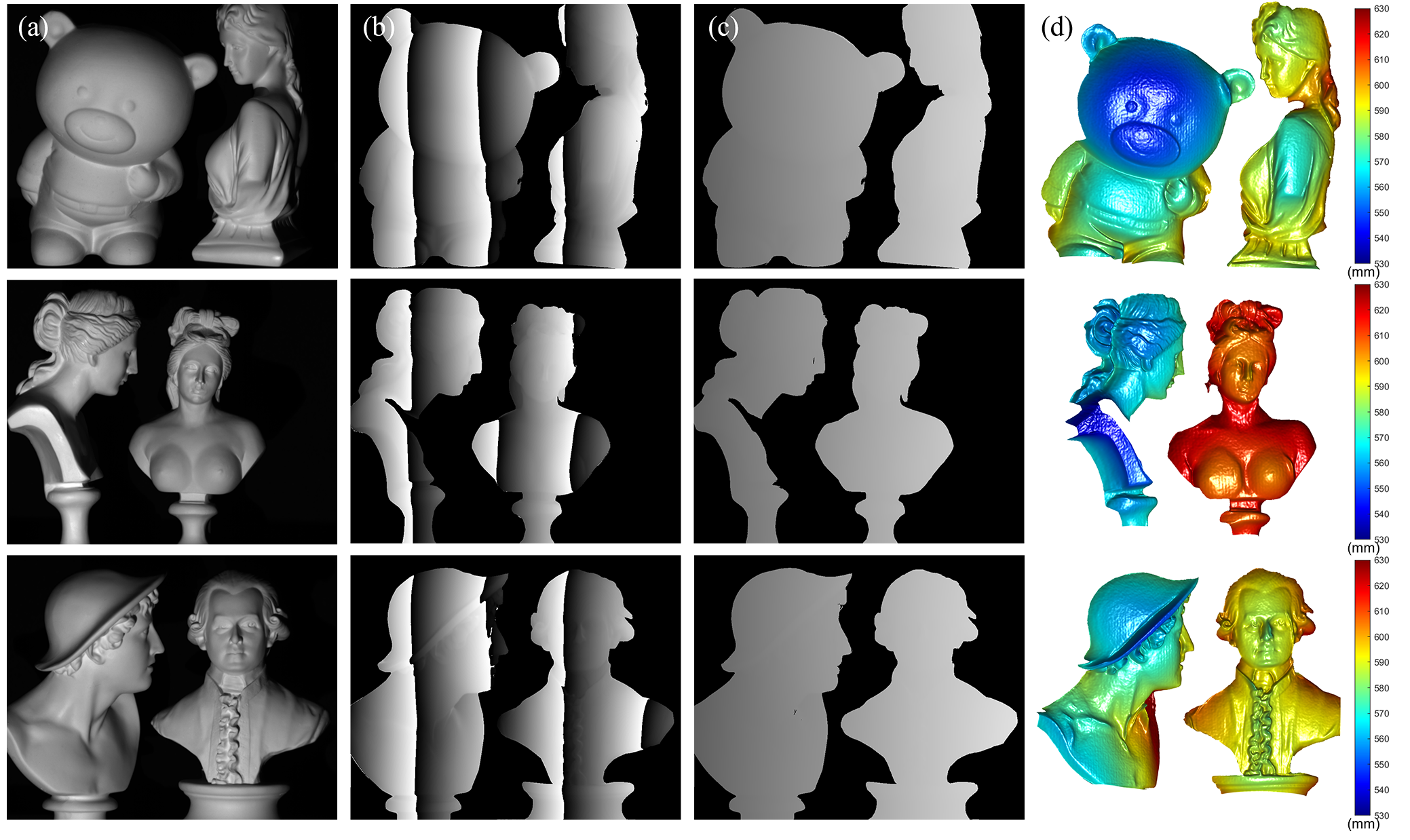}
\caption{Experimental results for multiple objects with arbitrary surface (a) Texture images of the statues, (b) Wrapped phase maps from the proposed method, (c) Unwrapped phase maps, (d) 3D reconstruction results of multiple objects with arbitrary surface. Each row corresponds to a different object, and each column presents the same type of result across all objects, enabling consistent visual comparison.}
\label{fig_13}
\end{figure}

We conducted an experiment to validate the reconstruction capability of the proposed method for arbitrary surface geometries. In this experiment, multiple isolated statues with significant depth variations and highly detailed regions, such as sculpted hair elements, were used, as shown in Fig.~\ref{fig_13}(a). Fig.~\ref{fig_13}(a) shows the texture image synthesized by combining the captured fringe patterns. Fig.~\ref{fig_13}(b) presents the wrapped phase map obtained using the least square method, and Fig.~\ref{fig_13}(c) shows the absolute phase map retrieved using the multi-wavelength phase unwrapping algorithm. The final 3D reconstruction result derived from the absolute phase and the polynomial regression model is illustrated in Fig.~\ref{fig_13}(d).

The results demonstrate that the proposed method can successfully retrieve a consistent absolute phase map across the entire FoV while simultaneously reconstructing spatially separated objects. Furthermore, fine surface details are well preserved in the reconstructed geometry. These results qualitatively confirm the capability of the proposed method to reconstruct objects with both spatial separation and complex surface geometries.

\subsection{Measurement uncertainty analysis}
\label{subsec_uncertainty}
To establish a rigorous characterization of the measurement accuracy, an uncertainty analysis of the reconstructed depth was conducted. The uncertainty analysis was conducted following approaches used in optical metrology systems employing structured-light measurements ~\cite{huang2023three,zheng2022universal}.

It should be noted that the planar validation experiment evaluates the internal consistency of the reconstruction process because the reference plane is obtained by fitting the measured data themselves. Therefore, the absolute accuracy and traceable uncertainty are evaluated using the sphere reference artifact.

The measurement result is modeled as the sum of the true value and independent error contributions as
\[
Z_m = Z + \sum_{i=1}^{5} \delta_i 
\] 
where $Z_m$ denotes the measured depth, $Z$ is the true depth, and $\delta_i$ represents the error component associated with each uncertainty source.

Thus, the combined standard uncertainty is expressed as
\begin{equation}
u_c(Z) =
\sqrt{
u_{\text{rep}}^2
+
u_{\text{repr}}^2
+
u_{\text{ref}}^2
+
u_{\text{cal}}^2
+
u_{\text{stage}}^2
}.
\label{eq:combined_uncertainty}
\end{equation}
The following subsections provide a detailed analysis of each uncertainty component.

\subsubsection{Standard Uncertainty Caused by Measurement Repeatability}
Under identical experimental conditions, the repeatability of the depth reconstruction was evaluated using 5 repeated measurements of the reference sphere, as described in Section~\ref{sec_sphere}. Conservatively, the largest STD of the sphere residual RMSE was $s = 0.018~\text{mm}$. The standard uncertainty introduced by measurement repeatability is calculated as follows:
\[
u_{\text{rep}} =
\sqrt{
\frac{1}{5(5-1)}
\sum_{i=1}^{5}(x_i-\bar{x})^2
}
=
\frac{s}{\sqrt{5}}
=
0.008~\text{mm}
\]

\subsubsection{Standard Uncertainty Caused by Measurement Reproducibility}

Measurement reproducibility was evaluated using the sphere experiment with 10 different poses described in Section~\ref{sec_sphere}. The STD of the pose-wise RMSE values was $s = 0.049~\text{mm}$. The standard uncertainty caused by reproducibility is calculated as follows:

\[
u_{\text{repr}} =
\sqrt{
\frac{1}{10(10-1)}
\sum_{j=1}^{10}(x_j-\bar{x})^2
}
=
\frac{s}{\sqrt{10}}
=
0.015~\text{mm}
\]

\subsubsection{Standard Uncertainty Caused by Reference Artifact}

The reference sphere used for validation has a manufacturer-specified diameter tolerance of $\pm 0.051~\text{mm}$. Considering a uniform distribution for the tolerance of the reference sphere, the corresponding standard uncertainty is 

\[
u_{\text{ref}} =
\frac{0.051}{\sqrt{3}}
=
0.029~\text{mm}.
\]

\subsubsection{Standard Uncertainty Caused by Calibration Residual}

The standard uncertainty associated with calibration was obtained from the pooled RMSE described in Section~\ref{sec_calib}. The calibration residual was statistically estimated using all calibration poses and valid pixels, yielding
\[
u_{\text{cal}} =
\sigma_{\text{cal}}
=
0.085~\text{mm}.
\]

\subsubsection{Standard Uncertainty Caused by Rotation Stage}

The effective phase-to-depth sensitivity is defined as
\[
S_{\mathrm{eff}}
=
\left\langle
\left|
\frac{\partial Z}{\partial \Phi}
\right|
\right\rangle .
\]
From the calibration data described in Section~\ref{sec_calib}, the effective sensitivity was estimated as $S_{\mathrm{eff}} = 39.304~\text{mm/rad}$ using all valid pixels across all poses.

In the proposed system, a phase period ($2\pi$) is generated over a total stage rotation of $\theta_h = 5^\circ$. Therefore, the phase sensitivity with respect to the stage angle can be approximated as
\[
\left|
\frac{\partial \Phi}{\partial \alpha}
\right|
\approx
\frac{2\pi}{\theta_h}.
\]

Since the effects of stage positioning accuracy and repeatability are implicitly reflected in the estimated calibration residual, only angular resolution is considered in the stage uncertainty budget. The angular resolution of the rotation stage (DPIN LSR-100L) was $\Delta\alpha = 0.004^\circ$ according to the manufacturer specifications. The standard uncertainty caused by the rotation stage is therefore expressed as
\[
u_{\text{stage}}
=
S_{\mathrm{eff}}
\left(
\frac{2\pi}{\theta_h}
\right)
\frac{\Delta\alpha}{\sqrt{12}} 
= 0.057~\text{mm}.
\]

The corresponding effective depth resolution of the system can be expressed as
\[
\Delta Z_{\text{eff}}
=
S_{\mathrm{eff}}
\left(
\frac{2\pi}{\theta_h}
\right)
\Delta\alpha
= 0.198~\text{mm}.
\]

\subsubsection{Combined and Expanded Uncertainty}

\begin{table}[t]
\centering
\caption{Summary of the uncertainty budget.}
\label{tab_uncertainty}
\begin{tabular}{l c c c}
\hline
Component & Symbol & Type & \makecell{Standard \\[3pt] uncertainty (mm)} \\
\hline
Measurement repeatability   & $u_{\text{rep}}$   & A & 0.008 \\
Measurement reproducibility & $u_{\text{repr}}$  & A & 0.015 \\
Reference sphere tolerance  & $u_{\text{ref}}$   & B & 0.029 \\
Calibration residual        & $u_{\text{cal}}$   & A & 0.085 \\
Rotation stage resolution   & $u_{\text{stage}}$ & B & 0.057 \\
\hline
Combined standard uncertainty & $u_c(Z)$ &  & 0.108 \\
Expanded uncertainty  & $U(Z)$   &  & 0.215 \\
\hline
\end{tabular}
\end{table}

Assuming independence among the individual uncertainty components, the combined standard uncertainty of the reconstructed depth is obtained from Eq.~(\ref{eq:combined_uncertainty}) as $u_c(Z) = 0.108~\text{mm}$.
Considering an approximate confidence level of 95\% (coverage factor of 2), the expanded uncertainty becomes $U(Z) = 0.215~\text{mm}$.
The complete uncertainty budget is summarized in Table~\ref{tab_uncertainty}.

The reported uncertainty applies to depth measurements obtained with the proposed system within the working depth range of 540–620 mm under the measurement conditions described in Sections~\ref{sec_calib} and~\ref{sec_sphere}. It should also be noted that the traceability of the reported measurement uncertainty is limited by the specification of the reference artifact used for validation. Consequently, the reported uncertainty represents the experimental measurement uncertainty of the proposed system under the present measurement configuration.

\section{Discussion}
Quantitative experiments on planar and spherical reference targets demonstrate that the proposed method achieves accurate reconstruction with stable measurement behavior. Across multiple poses, the proposed method achieves global reconstruction accuracy comparable to that of the SV-based reference method, as quantified by RMSE. Repeatability and reproducibility experiments further validate that the evaluated metrics exhibit limited dispersion across repeated acquisitions and varying measurement configurations. In addition, the measurement uncertainty analysis characterizes the measurement performance of the proposed method.

While the proposed method exhibits stable and accurate reconstruction, trade-offs with alternative approaches and systematic error sources remain to be analyzed.

\subsection{Trade-offs with Alternative Approaches}
Due to fundamentally different projection mechanisms and reconstruction principles, direct quantitative comparison with alternative approaches is not included, and a qualitative trade-off perspective is adopted.

Conventional DLP systems can temporally encode rich information and achieve high accuracy, but their FoV is inherently limited by unidirectional projection geometry. Metasurface-based projection supports omnidirectional illumination using a single static structured-light pattern, enabling dynamic scene measurement without mechanical actuation; however, it relies on learning-based reconstruction pipelines with high computational cost and limited geometric accuracy. Multi-view stereo--based omnidirectional approaches do not require active projection and allow flexible deployment, but tend to exhibit lower accuracy and strong sensitivity to surface texture and illumination. By comparison, the proposed phase-based method enables pixel-wise depth estimation while maintaining omnidirectional coverage.

\subsection{Analysis of Additional Error Sources}

\begin{table}[!t]
\caption{Regional RMSE comparison along the $u$-axis in the planar reproducibility experiment.}
\label{tab:rmse_uaxis_region}
\centering
\begin{tabular}{c|c|c}
\hline
Region along $u$-axis & Image area (\%) & Mean RMSE (mm) \\
\hline
Central region       & 30              & 0.060          \\
Outer region         & 30              & 0.071          \\
\hline
\end{tabular}
\end{table}

While the uncertainty analysis in Section ~\ref{subsec_uncertainty} quantifies the main contributors to the measurement uncertainty, several additional factors may also influence the reconstruction results. These effects are not modeled as independent uncertainty terms because their influence is implicitly reflected in the residual statistics used in the uncertainty budget. The following discussion briefly examines these additional sources of reconstruction error.

First, pattern distortion associated with the cylindrical projection geometry leads to spatially varying reconstruction errors, which increase toward the lateral regions along the $u$-axis. In the planar reproducibility experiment, the mean RMSE increases from 0.060~mm in the central 30\% of the $u$-range to 0.071~mm in the outer 30\%, defined as the left and right 15\% bands, as shown in Table~\ref{tab:rmse_uaxis_region}. Measurement consistency is further affected by system configuration errors, including rotational eccentricity of the projector and fabrication imperfections. Finally, phase noise caused by illumination variation and camera sensor noise introduces additional errors.

\subsection{Directions for Future Work}
Future work will prioritize reducing absolute reconstruction errors rather than further minimizing statistical dispersion metrics, as the experimental results suggest that error magnitude is the primary limitation to overall measurement performance. In particular, the increase in RMSE toward the lateral regions along the $u$-axis and the larger errors observed for spherical targets motivate focused improvements in distortion compensation and calibration accuracy. 

First, pixel-wise error compensation strategies will be investigated through quantitative analysis of the relationship between reference depth and pattern distortion, aiming to mitigate errors induced by the cylindrical projection geometry. Second, the phase--height relationship will be refined by comparing polynomial models of different orders and expanding the calibration dataset with a larger and more diverse set of reference poses in order to reduce calibration residual errors.

\section{Conclusion}

In this paper, we presented a 3D reconstruction method based on a cylindrical mechanical projector capable of omnidirectional fringe projection using phase-shifted patterns with two distinct spatial periods. The proposed design overcomes the limited projection coverage of conventional mechanical projector systems and enables absolute phase retrieval through a multi-wavelength unwrapping strategy. Furthermore, a quasi-calibration scheme allows 3D reconstruction to
be achieved within a single-camera, single-projector configuration. Experimental results, evaluated both quantitatively and qualitatively, demonstrate reconstruction performance comparable to that of an SV-based method. In addition, repeatability and reproducibility analyses based on multiple statistical metrics characterize the measurement performance and confirm consistent behavior across repeated acquisitions and varying measurement configurations. A measurement uncertainty analysis based on an uncertainty budget estimated a combined standard uncertainty of 0.108~mm and an expanded uncertainty of 0.215~mm for the reconstructed depth under the experimental conditions. These results validate the reliability and feasibility of the proposed method for omnidirectional 3D reconstruction in applications such as 3D telecommunication and industrial inspection, where wide-area dense measurement is required.

\section*{Acknowledgements}
This research was supported by the Technology Innovation Program(Project Name: Development of AI autonomous continuous production system technology for gas turbine blade maintenance and regeneration for power generation, Project Number: RS-2025-25447257, Contribution Rate: 50\%) funded By the Ministry of Trade, Industry and Resources(MOTIR, Korea), the Culture, Sports and Tourism R\&D Program through the Korea Creative Content Agency grant funded by the Ministry of Culture, Sports and Tourism in 2024 (Project Name: Global Talent for Generative AI Copyright Infringement and Copyright Theft, Project Number: RS-2024-00398413, Contribution Rate: 40\%), and the National Research Foundation of Korea(NRF) grant funded by the Korea government(MSIT) (Project Number: RS-2025-16072782, Contribution Rate: 10\%).

\bibliography{ref}
\bibliographystyle{IEEEtran}

\end{document}